\documentclass[10pt,twocolumn,letterpaper]{article}

\usepackage[pagenumbers]{cvpr} 

\usepackage{graphicx}
\usepackage{amsmath}
\usepackage{amssymb}
\usepackage{booktabs}
\usepackage{ifpdf}
\usepackage{algorithmic}
\usepackage{algorithm}
\usepackage{ulem}
\usepackage{CJKulem}
\usepackage{color}
\usepackage{times}
\usepackage{epsfig}
\usepackage{url}

\usepackage{amsmath}
\usepackage{amssymb}
\usepackage{diagbox}
\usepackage{multirow}
\usepackage{mathrsfs}
\usepackage{array}
\usepackage{float}
\usepackage{url}
\usepackage{array}
\usepackage{ulem}
\usepackage{soul}
\usepackage{algorithm}
\usepackage{algorithmic}
\usepackage[pagebackref,breaklinks,colorlinks]{hyperref}
\usepackage{hyperref}

\usepackage[capitalize]{cleveref}
\crefname{section}{Sec.}{Secs.}
\Crefname{section}{Section}{Sections}
\Crefname{table}{Table}{Tables}
\crefname{table}{Table}{Tabs.}

\makeatletter

\newcommand{\Rmnum}[1]{\expandafter\@slowromancap\romannumeral #1@}
\makeatother

\def\cvprPaperID{} 
\def\confName{}
\def\confYear{}

\begin{document}
\title{Open-World Pose Transfer via Sequential Test-Time Adaption}

\author{Junyang Chen\textsuperscript{1} \quad Xiaoyu Xian\textsuperscript{2} \quad Zhijing Yang\textsuperscript{1} \quad Tianshui Chen\textsuperscript{1} \quad Yongyi Lu\textsuperscript{1} \quad Yukai Shi\textsuperscript{1*} \\ Jinshan Pan\textsuperscript{3} \quad Liang Lin\textsuperscript{4}\\
\textsuperscript{1}Guangdong University of Technology \quad
\textsuperscript{2}CRRC Academy\\
\textsuperscript{3} Nanjing University of Science and Technology \quad
\textsuperscript{4} Sun Yat-sen University\\
{\tt\small \{jychen9811, tianshuichen, yylu1989, sdluran\}@gmail.com }\\ {\tt\small  \{yzhj, ykshi\}@gdut.edu.cn \quad xxy@crrc.tech \quad linliang@ieee.org}}

\twocolumn[{%
\renewcommand\twocolumn[1][]{#1}%
\maketitle
\begin{center}
\hsize=\textwidth 
\centering
\includegraphics[width=1.0\linewidth]{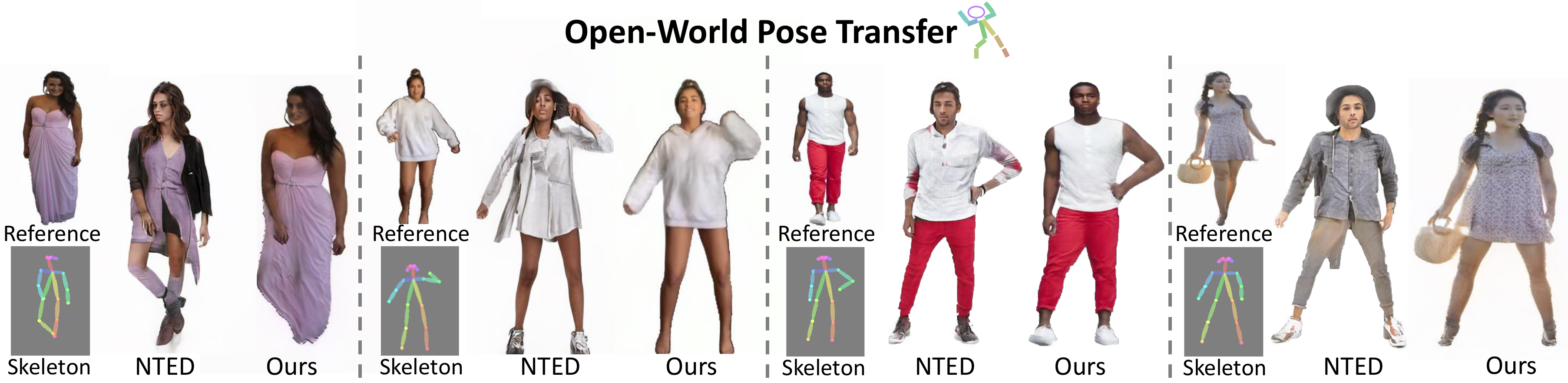}
\captionof{figure}{Visualization of open-world pose transfer (OWPT). With open-world references, it can be observed that typical pose transfer method (\emph{e.g.} NTED~\cite{Pose_one_4}) exhibits a twisty pattern. In this sense, we call for a solid model that can handle open-world instances beyond a specific dataset. }
\label{fig:OWPT}
\end{center}
}]


	    


\begin{abstract}
Pose transfer aims to transfer a given person into a specified posture, has recently attracted considerable attention. A typical pose transfer framework usually employs representative datasets to train a discriminative model, which is often violated by out-of-distribution (OOD) instances. Recently, test-time adaption (TTA) offers a feasible solution for OOD data by using a pre-trained model that learns essential features with self-supervision. However, those methods implicitly make an assumption that all test distributions have a unified signal that can be learned directly. In open-world conditions, the pose transfer task raises various independent signals: OOD appearance and skeleton, which need to be extracted and distributed in speciality. To address this point, we develop a SEquential Test-time Adaption (SETA). In the test-time phrase, SETA extracts and distributes external appearance texture by augmenting OOD data for self-supervised training. To make non-Euclidean similarity among different postures explicit, SETA uses the image representations derived from a person re-identification (Re-ID) model for similarity computation. By addressing implicit posture representation in the test-time sequentially, SETA greatly improves the generalization performance of current pose transfer models. In our experiment, we first show that pose transfer can be applied to open-world applications, including Tiktok reenactment and celebrity motion synthesis.
\end{abstract}

\section{Introduction}
Pose transfer aims at transforming a source person into a target posture, while maintaining the original appearance. Previously, some pose transfer works~\cite{Pose_one_1, Pose_one_2, Pose_one_3, Pose_one_4, Pose_one_5, Pose_one_6} achieve charming results on a specific dataset~\cite{liu2016deepfashion} by assuming the prior conditions of the target sample are very similar to the training sample. As shown in Fig.~\ref{fig:OWPT}, this assumption could be easily violated in practice due to the out-the-distribution (OOD) data from real-world applications. How to make those pose transfer models great again on OOD data is still a non-trivial challenge.


Recently, with the increasing attention of pose transfer, many datasets~\cite{SHHQ,DeepFashion2,Tiktok} are proposed for various contexts, which makes those datasets vastly different from each other across many domains. DeepFashion~\cite{liu2016deepfashion} contains more pretty consumers with fashion clothes, which makes SHHQ dataset~\cite{SHHQ} differ significantly from it in terms of clothing, age and posture. Tiktok~\cite{Tiktok} dataset has more dance routines that make itself different from DeepFashion in both posture and appearance. 

To show the discriminative characteristic of pose transfer datasets~\cite{SHHQ,DeepFashion2,Tiktok} with more quantitative evidence, we obtain the high-level feature by a person ReID~\cite{ReID_model} model and visualize them in Fig.~\ref{fig:motivation} (a). More specifically, the ReID model is trained on Market-1501~\cite{market_1501}, and the high-level features are fetched from the $1^{st}$ layer. As can be seen in Fig.~\ref{fig:motivation} (a), each dataset exhibits a distinct yet independent pattern, from which we realize such bias literally can be inherited by deep models. Once a discriminative model is trained on a specified dataset, the inherited bias affects its ability to process OOD data. Considering the application of pose transfer tasks in real-world scenarios, it is necessary to make pose transfer models overcome such bias by resolving the distribution shift between the source domain and test domains.

To alleviate the distribution shift problem, a family of methods~\cite{DA1,DA2,DA3,DA4,DA5,DA10,DA11,DA12} based on domain adaption (DA) have been proposed, which assume the target data are accessible during model adaption. Taking advantage of this privileged data, the generalization ability is significantly enhanced. However, user data is often considered private content, which makes it unfeasible for upload, annotation and re-training. Domain generalization~\cite{DG1,DG2,DG3,DG4,DG5,DG6} (DG) addresses domain shift without pre-fetching OOD data. It extends the diversity of datasets to learn more generalized features. Nevertheless, constructing an expensive dataset does not always hold in practice, new cases always appear to cause new trouble. Recently, test-time adaption (TTA) methods~\cite{Tent, TTT} blur the boundary between DA and DG by assuming OOD data can be used in local device without annotation. As suggested in \cite{TTT}, in TTA, a pre-trained model learns essential feature representations from OOD data with self-supervision~\cite{gandelsman2022test}. In this sense, TTA~\cite{TTT,Tent} methods simply optimize model for the test distribution straightway. This flexible learning paradigm makes strong adaptability toward OOD data. However, those methods implicitly make an assumption that \emph{all test distributions have a unified signal that can be learned directly.}

In open-world conditions, the pose transfer task raises various yet non-trivial signals: out-the-distribution (OOD) appearance and skeleton. This means there are various domain knowledge required to be extracted in speciality. As shown in Fig.~\ref{fig:motivation} (b), we train NTED~\cite{Pose_one_4} with DeepFashion dataset, and apply the trained model on DeepFashion2, SHHQ and Tiktok datasets for inference. As can be seen in Fig.~\ref{fig:motivation} (b), the generated results easily show a twisty pattern with OOD appearance/skeleton. However, typical TTA methods were not designed to learn disentangled signals individually and originally. Therefore, we develop a SEquential Test-time Adaption (SETA) for independent signals adaption. Meanwhile, we develop an appearance adapter for external appearance texture extraction and distribution. Since the postures derived from OOD skeletons are non-Euclidean, we employ a person Re-ID~\cite{ReID_model} to extract the representations of each 
posture to compute the consistency over motion. SETA significantly improves the generalization performance toward OOD instances by fetching independent OOD signals sequentially. The contributions are summarized as follows:

\begin{figure*}[ht]
	\begin{center}
	    
		\includegraphics[width=0.95\linewidth,height=0.35\linewidth]{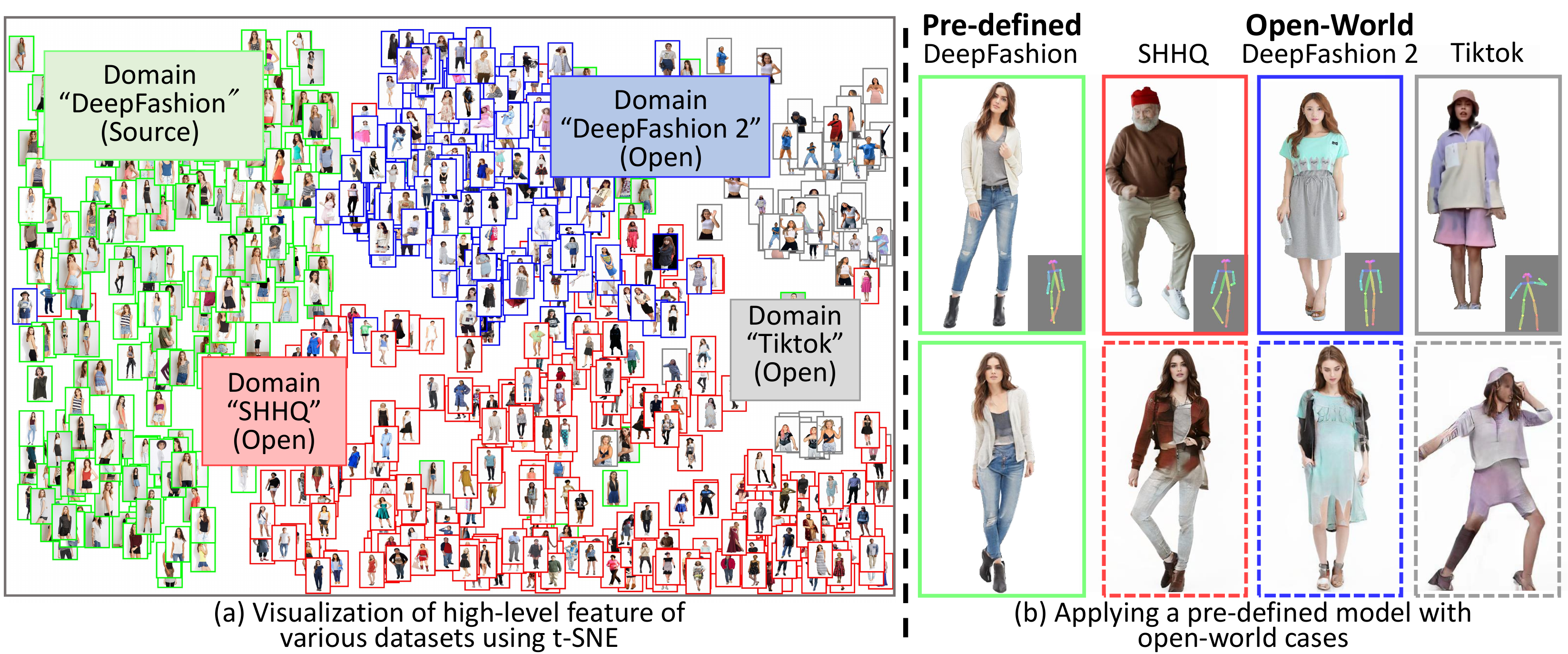}
	\end{center}
 	\vspace{-5mm}
	\caption{(a) To investigate the domain gap between source and OOD datasets, we obtain the high-level feature by a person ReID model~\cite{ReID_model} and use t-SNE for visualization. (b) The domain generalization of NTED~\cite{Pose_one_4}. Typical pose transfer model performs reasonably well on the source domain, however, the generated results could be easily violated with OOD input. We first propose an Open-World Pose Transfer (OWPT) framework to investigate the domain generalization of a pre-defined model toward OOD appearance and skeleton. }
	\label{fig:motivation}
	\vspace{-6mm}
\end{figure*}



(1)    Test-time adaption (TTA) is renewed for a various domain scene. In our model, multiple OOD domain knowledge ( \emph{e.g.} appearance, posture ) are SEquentially learned and distributed in Test-Time Adaption (SETA).  

(2)    To learn implicit appearance representation among OOD postures, SETA cleverly employs the image representations derived from a person re-identification (Re-ID) model to obtain non-Euclidean consistency over motion.

(3)    Pose transfer is further extended for an open-world environment for the first time. For the first time, we conduct pose transfer into open-world applications, including Tiktok reenactment and celebrity motion synthesis.

\vspace{-2mm}
\section{Related Work}

\paragraph{Test-Time Adaption.} Test-Time Adaption (TTA) aims to use the existing model to quickly adapt to OOD data during the test stage. Recently, some TTA methods~\cite{Tent, TTT, Lilun1, Lilun2, Lilun3, Lilun4} have been proposed for model generalization to OOD data. Sun \emph{et al}.~\cite{TTT} propose to apply the self-supervised proxy task to update model parameters on target data. Wang \emph{et al}.~\cite{Tent} introduce a entropy minimization method to optimize the parameters of the batch normalization layers. As the promising application prospect, TTA has been extended to several tasks~\cite{Deblurring, Dehazing, Super_resolution, RL, MM_TTA}. Chi \emph{et al}.~\cite{Deblurring} propose a meta-auxiliary learning paradigm for fast updating model parameters in dynamic scene deblurring task. Liu \emph{et al}.~\cite{Dehazing} update model parameters with the self-training signals from the proposed self-reconstruction method. Shin \emph{et al}.~\cite{MM_TTA} introduce cross-modal pseudo labels as self-training signals. These previous works always focus on single kind of self-training signal. However, when facing various OOD domain knowledge, the typical paradigm of TTA needs to be re-examined. In comparison to the aforementioned works, we need to provide various self-training signals in open world pose transfer. Hence, we develop a Sequential Test-Time Adaption for learning multiple domain knowledge with various self-training signals.

\vspace{-4mm}

\paragraph{Pose Transfer.} Pose Transfer has been an attractive topic in the image synthesis community since Ma \emph{et al}.~\cite{Pose_1} proposed. Up to now, many pipelines have been proposed, which can be classified as multi-stage approaches~\cite{Pose_multi_1, Pose_multi_2, Pose_multi_3, Pose_multi_4} and one-stage methods~\cite{Pose_one_1, Pose_one_2, Pose_one_3, Pose_one_4, Pose_one_5, Pose_one_6}. The former presents a coarse-to-fine structure which utilizes coarse shape or foreground masks to ensure generalization on arbitrary poses. However, these methods are not efficient at inference and require additional computing power. ADGAN~\cite{Pose_one_2} ameliorates this issue by extracting expressive textures vectors from different semantic entities to synthesize the target image. CASD~\cite{Pose_one_3} introduces attention-based methods to distribute the semantic vectors to the target poses. Compared to these parser-based methods, NTED~\cite{Pose_one_4} applies sparse attention-based operation to extract the semantic textures without the assistance of the external parser. Although these methods have been validated well on the DeepFashion dataset~\cite{liu2016deepfashion}, there is no relevant research to extend these pre-trained models to OOD dataset. Our work first explores the performance of these models on OOD data. However, the performance still remains a large gap between photo-realistic image due to the pre-trained models overfitting on DeepFashion dataset~\cite{liu2016deepfashion}.

\vspace{-4mm}
\section{Preliminaries}

\subsection{Pose Transfer}
Given a reference image, traditional pose transfer methods aim at synthesizing high fidelity images with different poses. Most pose transfer approaches adopt a similar pipeline: a \emph{texture encoder} extracts appearance characters from the reference images, a \emph{skeleton encoder} describes the semantic distribution from the target poses. Finally, a \emph{generator} is used to produce high fidelity images via transferring the appearance texture to the target poses under the semantic distribution guidance. 

\subsection{Bottlenecks}
Traditional pose transfer approaches have the ability to extract and distribute the human appearance texture within a fixed dataset. However, as shown in Fig.~\ref{fig:motivation}, this ability is limited by the domain gap. In the inference stage, since the training data and test data are drawn from different distribution, even a minor differences turn out to weaken state-of-the-art approaches. Thus, \emph{`How to fetch desired signals from OOD samples'} is crucial to pave the way for open-world pose transfer task.

\section{Proposed Method}
Given a discriminative model $f_{\theta}$ trained on a representative dataset~\cite{liu2016deepfashion}, our goal is to generate the realistic images on out-of-distributions (OOD) data. In this section, we introduce an effective solution to learn the disentangled OOD signals (\emph{i.e.} OOD apearance signals, OOD skeleton signals) sequentially. First, we update the pre-trained parameters ${\theta}$ via OOD appearance signals, which are fetched from OOD data and its augmentation. Then, we introduce the ReID model to obtain the consistency over motion during pose transformation. Finally, we fetch OOD skeleton signals from the consistency over motion to update the previously adapted parameters.


\begin{figure*}[t]
     \vspace{-1mm}
	\begin{center}
		\includegraphics[width=0.95\linewidth,height=0.33\linewidth]{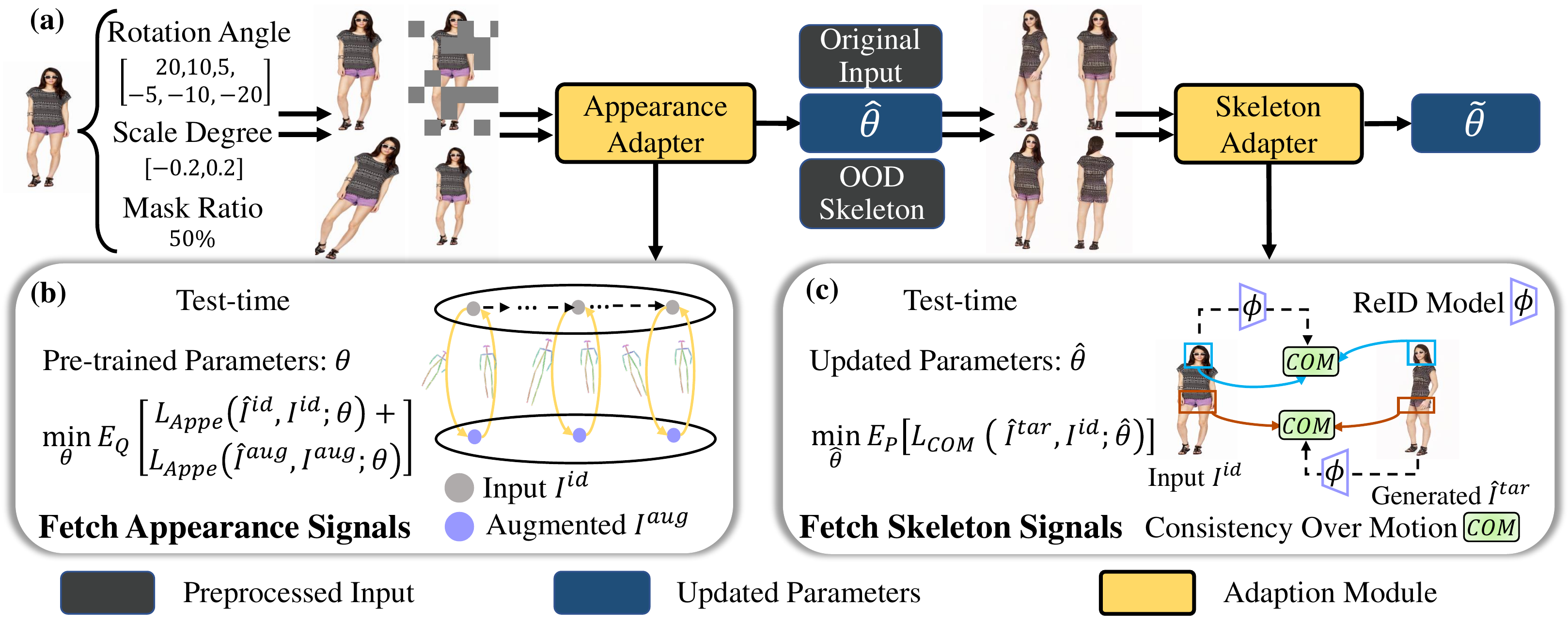}
	\end{center}
 	\vspace{-5mm}
	\caption{(a) Overview of Sequential Test-Time Adaption. (b) We optimize the pre-trained parameters $\theta$ by external appearance signals, which fetched from OOD data with augmentation, in the test stage. Note that $\hat{I}^{id} = f_{\theta}(I^{aug},P^{id})$ and $\hat{I}^{aug} = f_{\theta}(I^{id},P^{aug})$. $P^{id}$ and $P^{aug}$ are the skeleton of $I^{id}$ and $I^{aug}$. (c) We introduce arbitrary OOD skeletons to generate the pose transfer images with the updated parameters $\hat{\theta}$. A person ReID model~\cite{ReID_model} is used to obtain the consistency over motion between the input image and generated images. Then, we fetch the implicit posture representation from the consistency to optimize $\hat{\theta}$ in the test-time adaption stage for OOD skeletons. }
	\label{fig:pipeline}
 	\vspace{-5mm}
\end{figure*}



\subsection{Open Appearance}

Inspired by recent progress in TTA, we develop an \textbf{\emph{appearance adapter}} to learn essential knowledge from OOD appearance sequentially.

\vspace{-4mm}

\paragraph{Fetch Appearance Signals.}
In test-time phrase, given a person image $I^{id}$, we apply the appearance adapter to learn OOD appearance signals. First, we use data augmentation to generate samples for self-supervised training. Each augmentation can transform data stochastically with some internal parameters (\emph{e.g.} rotation angle, scale degree, mask ratio). The augmented samples are denoted as $I^{aug}$. As shown in Fig.~\ref{fig:pipeline} (b), given a pair of training images $(I^{id}, I^{aug})$, a pre-trained model $f_\theta$ with parameter $\theta$, we update $f_\theta$ to learn the signals from OOD appearance domain based on the loss $L_{Appe}$, which includes reconstruction loss $L_{rec}$, perceptual loss~\cite{VGGloss} $L_{perc}$ and 
attention loss~\cite{Pose_one_4} $L_{att}$. Thus, the appearance adaptive loss is given as follows:  
\begin{equation}
    \begin{split}
        L_{Appe}=L_{rec} + L_{perc} + L_{att}.
    \end{split}
\end{equation}
where $L_{rec}$ is formulated as the L1 distance, $L_{perc}$ is computed on the VGG layers, and $L_{att}$ is used to calculate L1 distance at each operation layer.

In test-time phrase, we optimize the self-supervised loss $L_{Appe}$ over OOD data and its derivatives that drawn from the test distribution $Q$, which is defined as follows:
\begin{equation}
    \begin{split}
    \hat{\theta} = &\mathop{\arg\min}_{\theta} \mathbb{E}_Q\begin{bmatrix}
        L_{Appe}(\hat{I}^{id}, I^{id}; \theta)+\\
        L_{Appe}(\hat{I}^{aug}, I^{aug}; \theta)
        \end{bmatrix}.
    \end{split}
\end{equation}
where $\hat{I}^{id} = f_{\theta}(I^{aug},P^{id})$ and $\hat{I}^{aug} = f_{\theta}(I^{id},P^{aug})$, $P^{id}$ and $P^{aug}$ are the skeleton of $I^{id}$ and $I^{aug}$. $\theta$ and $\hat{\theta}$ indicate the pre-trained model parameters and the adapted parameters learned from OOD appearance signals, and $E_Q$ is evaluated on an OOD appearance distribution $Q$. 





\paragraph{Open Appearance Deployment.} The updated model $f_{\hat{\theta}}$ has been learned specifically to facilitate adaptation to OOD appearance domain. In inference stage, the appearance texture of OOD data could be extracted and then distributed according to the semantic distribution of skeleton via $f_{\hat{\theta}}$. 


\subsection{Open Skeleton}

\paragraph{Fetch Skeleton Signals.}
To learn the implicit posture representation in pose transformation, we develop a \textbf{\emph{skeleton adapter}} to fetch OOD skeleton signals from the consistency over motion. As shown in Fig. \ref{fig:pipeline} (c), non-Euclidean consistency exists between the original and transferred person image.



\begin{table*}[t]
	\centering
	\vspace{2mm}
    \caption{Evaluation results on Open-World Pose Transfer (OWPT). }
    \vspace{0mm}
    	   \setlength{\tabcolsep}{2mm}
	   {
\begin{tabular}{ccccccccccccc}
\toprule[1pt]
\multicolumn{1}{c}{\multirow{2}{*}{Method}} & \multicolumn{1}{c}{\multirow{2}{*}{OWPT}} & \multicolumn{3}{c}{SHHQ} &  & \multicolumn{3}{c}{DeepFashion 2} &  & \multicolumn{3}{c}{Tiktok} \\ \cline{3-5} \cline{7-9} \cline{11-13} 
\multicolumn{2}{c}{}                        & SSIM~{\color{red}$\uparrow$}   & LPIPS~{\color{red}$\downarrow$}   & FID~{\color{red}$\downarrow$}   &  & SSIM~{\color{red}$\uparrow$}      & LPIPS~{\color{red}$\downarrow$}     & FID~{\color{red}$\downarrow$}     &  & SSIM~{\color{red}$\uparrow$}    & LPIPS~{\color{red}$\downarrow$}   & FID~{\color{red}$\downarrow$}   \\ \hline
\multirow{2}{*}{ADGAN}          & w/o SETA           &0.586        &0.425         &73.04       &  &0.608           &0.444           &75.53         &  &0.657         &  0.291        &77.14       \\
                                & w/ SETA          &\textbf{0.901}         & \textbf{0.079}       &\textbf{35.67}       &  &\textbf{0.885}           & \textbf{0.146}          &\textbf{65.09}         &  &\textbf{0.834}         & \textbf{0.133}         &\textbf{68.35}       \\ \hline
\multirow{2}{*}{CASD}           & w/o SETA          & 0.728        & 0.198         & 36.09      &   &  0.702         &0.268           & 45.09         &  &0.671         & 0.277         & 63.33      \\
                                & w/ SETA           &\textbf{0.933}        &\textbf{0.042}         &\textbf{18.66}       &  & \textbf{0.916}          &\textbf{0.056}          & \textbf{32.15}         &  &\textbf{0.819}         & \textbf{0.109}         & \textbf{55.21}      \\ \hline
\multirow{2}{*}{NTED}           & w/o SETA         & 0.723        & 0.211         &40.25       &  &0.686           & 0.284          &53.457         &  &0.678         &0.259          &72.68        \\
                                & w/ SETA          & \textbf{0.890}       &\textbf{0.049}         &\textbf{19.07}       &  &\textbf{0.859}           & \textbf{0.089}         & \textbf{33.74}        &  &\textbf{0.837}         &\textbf{0.097}          &\textbf{35.56}       \\
\bottomrule[1pt]

\label{tab:quantitative}
\end{tabular}}
\vspace{-8mm}
\end{table*}

\paragraph{Global Consistency over Motion.} We fetch OOD skeleton signals from the consistency over motion. First, we use the model $f_{\hat{\theta}}$ to generate the pose transfer image $\hat{I}^{tar}$ from the reference image $I^{id}$ and target skeleton $P^{tar}$. \emph{A person re-identification model has potential to search the images of same person with different posture}, with the help of that, we obtain the consistency over motion. Specifically, we apply a ReID model~\cite{ReID_model} on the reference person image and the generated image to obtain features for similarity computation~\cite{ReID_loss}:

\begin{equation}
L_{Content}=\sum_{t}\|\phi_t(\hat{I}^{tar})-\phi_t(I^{id})\|_2.
\label{REID}
\end{equation}
where $\phi_t$ represents the $t$-th layer of ReID model~\cite{ReID_model}. 

\vspace{-4mm}

\paragraph{Local Consistency over Motion.} In addition, we also enforce the local correspondence between each body part of the generated image and the reference image. However, the regions of the same body part in different poses usually have different sizes and shapes, which prevents us from computing loss in Euclidean space (\emph{e.g.} SSIM, L2 and perceptual loss). Inspired by \cite{Gram}, we use the Gram matrix to calculate the local similarity loss, which is not restricted to Euclidean space. Thus, we compute the Gram-matrix similarity using ReID's feature within each body-part as:
\begin{equation}
L_{GRAM}=\sum_{q}\|G(\hat{M}_q\odot\phi_t(\hat{I}^{tar}))-G({M}_q\odot\phi_t(I^{id}))\|_2.
\end{equation}
where $M_q$ and $\hat{M}_q$ are the human parsing results of $I^{id}$ and $\hat{I}^{tar}$ estimated by~\cite{Parsing}, $G$ is the Gram matrix, $\odot$ denotes the element-wise multiplication. We use the features from the first layer (\emph{i.e.} $t = 1$) empirically. Thus, we calculate the consistency loss over motion via the global and local correspondence as:
\begin{equation}
L_{COM}=L_{Content} + L_{GRAM}.
\label{motion_consistency}
\end{equation}



Note that $L_{COM}$ is a function for updating model parameters $\hat{\theta}$, which is used to learn external skeleton signals during pose transformation. In test-time phase, we can optimize $L_{COM}$ over OOD skeleton drawn from a test distribution $P$, which is defined as follows:
\begin{equation}
    \begin{split}
    \tilde{\theta} = &\mathop{\arg\min}_{\hat{\theta}} \mathbb{E}_P\begin{bmatrix}
        L_{COM}(\hat{I}^{tar}, I^{id}; \hat{\theta})
        \end{bmatrix}.
    \end{split}
\end{equation}
where $\hat{I}^{tar} = f_{\hat{\theta}}(I^{id},P^{tar})$. $P^{tar}$ is OOD skeletons. $\tilde{\theta}$ indicates the updated parameters learned from OOD skeleton signals. $E_P$ is evaluated on an OOD skeleton distribution $P$. 




Ideally, we update the pre-trained parameters $\theta$ via the sequential OOD signals (\emph{i.e.} appearance$\rightarrow$skeleton). The sequential optimization is given as follows:
\begin{equation}
    \begin{split}
    \hat{\theta} \gets \theta-\alpha\nabla_\theta L_{Appe}(\hat{I}^{id}, \hat{I}^{aug}, I^{id}, I^{aug}; \theta)
    \end{split}
\end{equation}
\begin{equation}
    \begin{split}
    \tilde{\theta} \gets \hat{\theta}-\beta{\nabla_{\hat{\theta}}}L_{COM}(\hat{I}^{tar},I^{id};\hat{\theta})
    \end{split}
\end{equation}
where $\alpha$ and $\beta$ are the adaptation learning rates. We present a summarization of SETA in Algorithm~\ref{alg:Framwork}.


\renewcommand{\algorithmicrequire}{\textbf{Input:}}  
\renewcommand{\algorithmicensure}{\textbf{Output:}} 

\begin{figure*}[t]
	\begin{center}
		\includegraphics[width=0.95\linewidth]{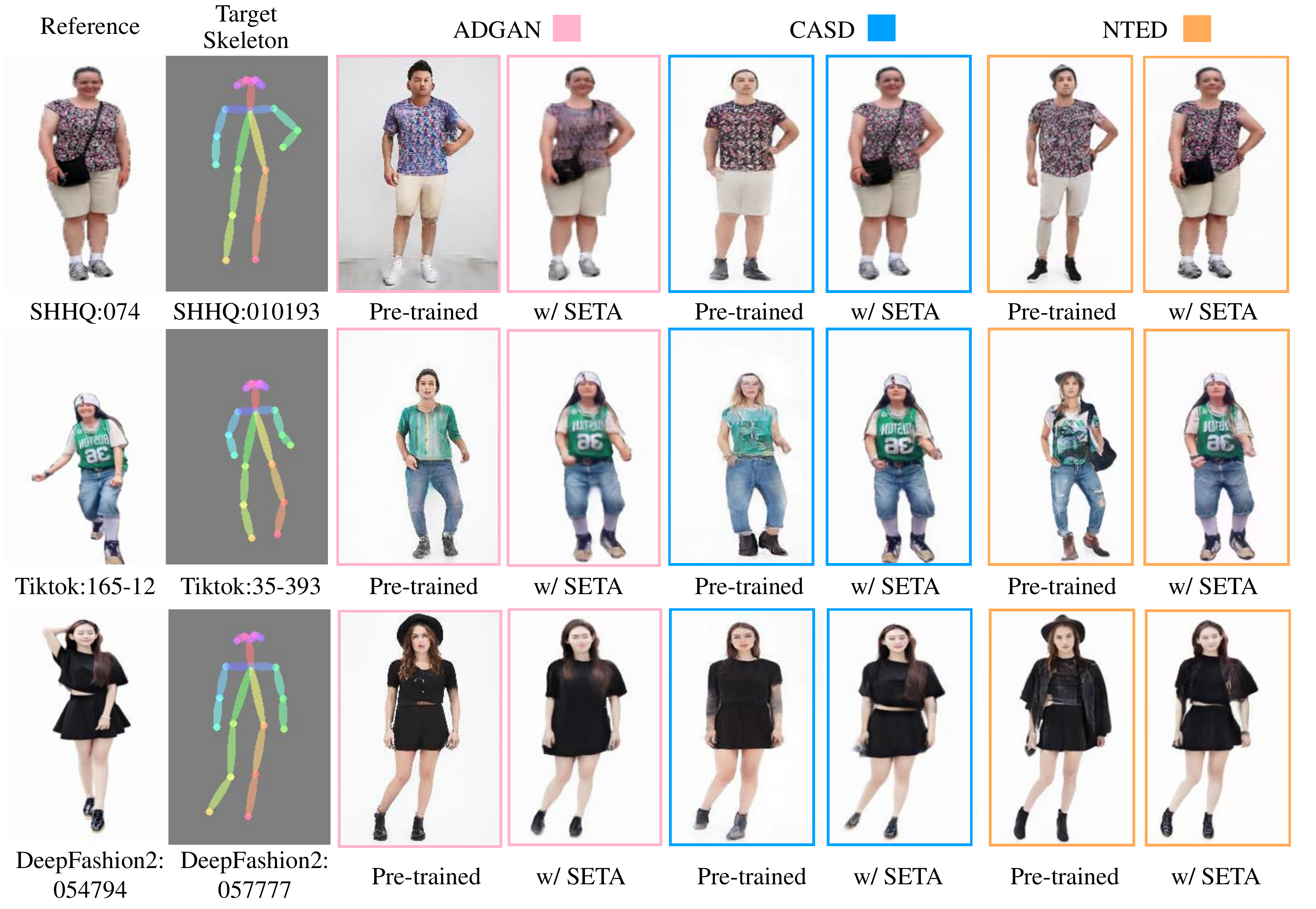}
	\end{center}
 	\vspace{-5mm}
	\caption{Qualitative comparison of using our proposed method on different datasets. Pose transfer frameworks generate more realistic on ODD references with SETA.}
	\label{fig:quanlitation}
	\vspace{-6mm}
\end{figure*}

\section{Experiments}

In this section, we describe our experimental setups and evaluate our proposed methods on benchmark datasets with various pre-trained models. We also apply our methods to other human generation tasks, such as celebrity motion synthesis and skeleton-driven tiktok reenactment, to show the potential extensibility of our approaches.

\begin{algorithm}[t]
\caption{SETA algorithm.}
\label{alg:Framwork}
\begin{algorithmic}[1] 
    \REQUIRE $\alpha$, $\beta$:  learning rates\\
    \REQUIRE $I^{id}$, $P^{id}$: OOD person image and skeleton \\
    \REQUIRE $P^{tar}_v$: target skeleton \\
    \REQUIRE  $\theta$: pre-trained model
    
    \STATE Sample an batch of OOD data in $\{I^{id}, P^{tar}_v\}^K_{v=1}$;
    \STATE Augment $\left \{ I^{id},P^{id} \right \}$ to generate set $\left \{ I^{aug},P^{aug} \right \}$;
    \STATE Compute $L_{Appe}(I^{id},P^{id},I^{aug},P^{aug})$;
    \FOR { $\left \{ I^{aug}_i,P^{aug}_i \right \}$ in $ \left \{ I^{aug},P^{aug} \right \}$}
    \STATE Generate fake images:
    \STATE $\hat{I}^{id} = f_{\theta}(I^{aug}_i,P^{id})$ and $\hat{I}^{aug}_i = f_{\theta}(I^{id},P^{aug}_i)$;
    \STATE Update parameters with gradient descent:\\ 
    $\hat{\theta} \gets \theta-\alpha\nabla_\theta L_{Appe}(\hat{I}^{id}, \hat{I}^{aug}_i, I^{id}, I^{aug}_i; \theta)$;
    \ENDFOR
    \WHILE {$v \le  K$}
        \STATE Generate fake images: $\hat{I}^{tar}_v = f_{\hat{\theta}}(I^{id},P^{tar}_v)$;
        \STATE Compute $L_{COM}(\hat{I}^{tar}_v, I^{id})$;
        \STATE Update parameters with gradient descent:\\ $\tilde{\theta} \gets \hat{\theta}-\beta{\nabla_{\hat{\theta}}}L_{COM}(\hat{I}^{tar}_v,I^{id};\hat{\theta})$;
    
    \ENDWHILE
    \ENSURE ~~\\ 
        Updated model parameter $\tilde{\theta}$.
\end{algorithmic}
\end{algorithm}

\subsection{Implementation Details}

In our experiments, the pre-trained models from NTED, CASD and ADGAN are trained on the In-shop Clothes Retrieval Benchmark of the DeepFashion dataset~\cite{liu2016deepfashion} (\emph{e.g.} images of the same person are paired). During test-time adaption stage, the rotation angle is set to \{20,10,5,-5,-10,-20\}, scale degrees is set to [-0.2,0.2], and the mask ratio is set to 50\%. For OOD appearance adaptation step, we perform 30 training iterations. Then, we perform 5 iterations with skeleton domain adapter with OOD skeletons. Both two stages use the initial learning rate as $2 \times 10^{-3}$ of all networks except the NTED~\cite{Pose_one_4}, which is set to be $1 \times 10^{-3}$. The Adam solver is used for both adaption stage with hyper-parameter $\beta_1=0.5$, and $\beta_2=0.99$. Note that $\beta_1=0$ in NTED. All the experiments are conducted on Nvidia V100 GPUs.


\subsection{Evaluation Datasets and Metrics}
Under the Open-World Pose Transfer (OWPT) setting, we employ SHHQ~\cite{SHHQ}, DeepFashion 2~\cite{DeepFashion2} and Tiktok~\cite{Tiktok} datasets for evaluation. SHHQ~\cite{SHHQ} is currently the largest dataset of human whole body, which consists of various appearance and poses styles person images. DeepFashion 2~\cite{DeepFashion2} contains lots of Asians in comprehensive fashion outfits. Tiktok~\cite{Tiktok} dataset consists of dance videos that capture a single person performing dance moves.  We use 30 videos from Tiktok dataset, 2536 images from SHHQ, and 1557 images from DeepFashion 2 as the evaluation datasets. For all images, we process them into image resolution of 256~$\times$~176. We adopt SSIM\cite{SSIM}, LPIPS\cite{Lpips} and FID\cite{FID} (Frechet Inception Distance) as the evaluation metrics.

\subsection{Comparisons}

\paragraph{Quantitative Comparison.} Under the OWPT setting, the performance of baseline approaches with our proposed sequential test-time adaption is reported under the term `w/ SETA'. As shown in Table~\ref{tab:quantitative}, our methods consistently outperform the existing approaches. The quantitative results on SSIM and LPIPS demonstrate that our approach obtains better image similarity toward OOD data. In addition, our approaches outperform the baseline methods with lower FID scores, which indicate better-quality images are shown by SETA.

\begin{figure*}[t]
\centering
\includegraphics[width=0.95\linewidth,height=0.28\linewidth]{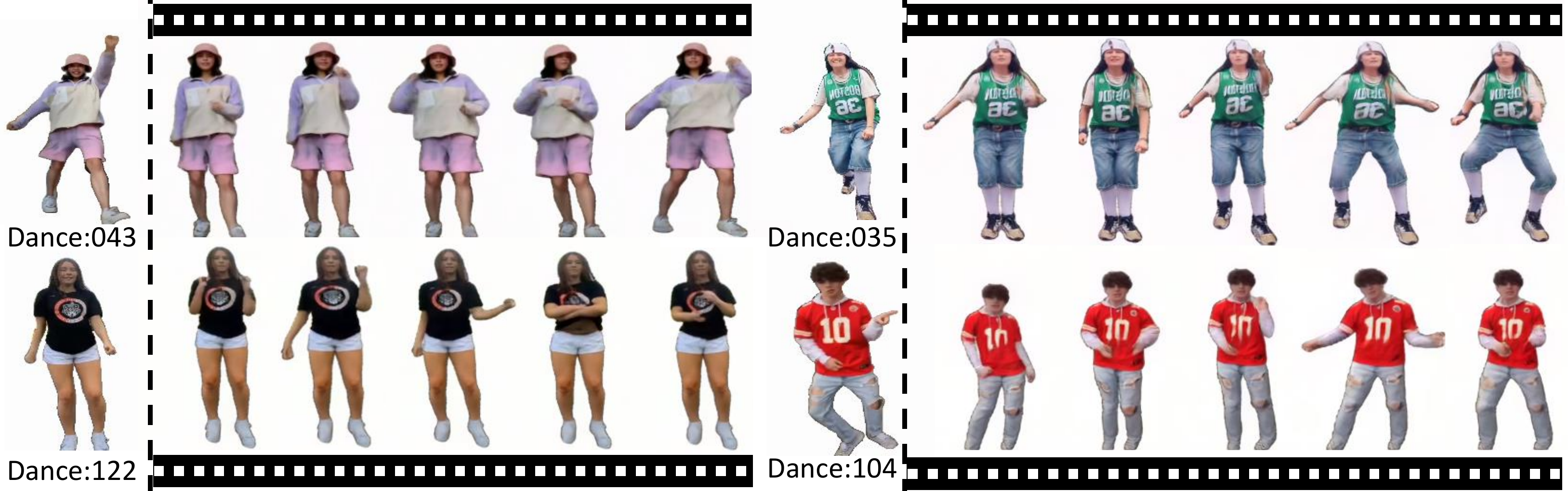}
\caption{Skeleton-driven results on the Tiktok dataset. Given target skeletons, our method can generate realistic dance sequences.}
\label{fig:video}
\vspace{-3mm}
\end{figure*}

\begin{figure*}[t]
\centering
\includegraphics[width=0.95\linewidth,height=0.23\linewidth]{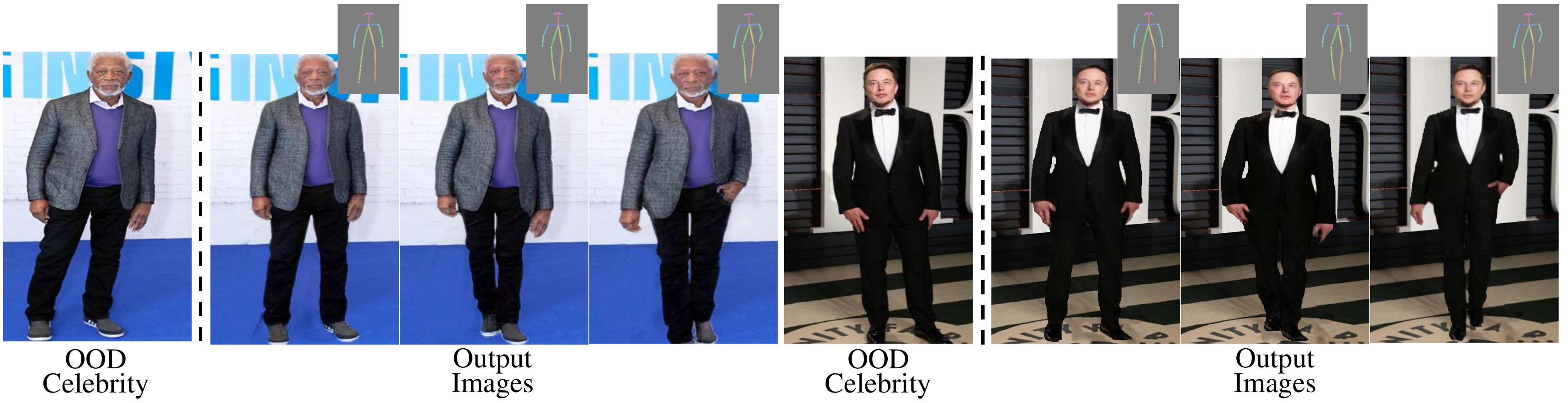}
\vspace{-2mm}
\caption{Examples of celebrity motion synthesis. Our algorithm transfer Morgan Freeman and Elon Musk into desired postures.}
\label{fig:celebrity}
\vspace{-5mm}
\end{figure*}

\vspace{-4mm}
\paragraph{Qualitative Comparison.} To further validate the proposed SEquntial Test-Time Adaption method under OWPT setting, we perform visual comparison of our method with recent proposed pose transfer methods in Fig.~\ref{fig:quanlitation}, including ADGAN~\cite{Pose_one_2}, CASD~\cite{Pose_one_3} and NTED~\cite{Pose_one_4}. As shown in the first and second rows of Fig.~\ref{fig:quanlitation}, when the appearance texture and body shape of reference person are different from the source dataset, these attributes of the generated results do not match to each other. Therefore, pre-trained models fail to preserve appearance texture, which demonstrate the challenge from OOD data. In the last row of Fig.~\ref{fig:quanlitation}, since the race appearance of Deepfashion 2 is different from the source dataset, images generated by baseline models fail to keep consistent appearances. This reveals pre-trained models could not be applied to the unknown data as they are limited by the prior knowledge inherit from source dataset.

In comparison, the generated results `w/ SETA' demonstrate the effectiveness of SEquential Test-Time Adaption under OWPT setting. As shown in Fig.~\ref{fig:quanlitation}, the adapted model has learned the OOD appearance signals, which could preserve the gender and clothes texture well. Benefited from the skeleton adapter, the updated model is able to distribute the texture to the target skeleton reasonably. Since we do not need additional pair labels of the input image during TTA stage, SETA can adaptively extend to various single human images of OOD data.





\vspace{-3mm}
\subsection{Skeleton-driven Tiktok Reenactment}
\vspace{-2mm} In this subsection, we show that our model can generate coherent single-person dance videos with delightful visual performance. We first extract appearance signals and skeleton signals from sequential motions with various poses of source video by our SETA. Then, we transform a reference person image into sequence dance skeletons. We conduct experiment on the Tiktok dataset~\cite{Tiktok} and 30 videos are used for testing. As shown in the second row of Fig.~\ref{fig:quanlitation}, state-of-the-art methods~\cite{Pose_one_4, Pose_one_3, Pose_one_2} fail to keep the source identity, leading inability to generate the realism of the produced videos. In contrast, combined with SETA, the pose transfer model~\cite{Pose_one_4, Pose_one_3, Pose_one_2} can generate realistic results along with accurate movements while still preserving the source identity. Qualitative reenactments results generated by SETA are shown in Fig.~\ref{fig:video}, which synthesis delightful visual quality even with complex poses input. We provide more skeleton-driven tiktok~\cite{Tiktok} driven reenactment samples in Fig.~\ref{fig:video_comp}.

\subsection{Open-World Celebrity Motion Synthesis} In this subsection, we aim to generate high-resolution celebrity image with arbitrary poses. We apply SETA to learn the celebrity's appearance and generate different pose views with $512 \times 352$ resolution. Experiments conducted on Morgan Freeman and Elon Reeve Musk are provided in Fig.~\ref{fig:celebrity}. It can be seen that our SETA can generate realistic results with well-preserved source identity details, such as clothes, human body parts and facial expression. To this end, SETA is able to synthesize high-resolution realistic results for celebrity motion application. More celebrity motion synthesis samples are shown in Fig.~\ref{fig:celebrity_1} and Fig.~\ref{fig:celebrity_2}.


\subsection{Analysis and Discussion}
We perform ablation studies to further investigate various aspects of the proposed approach. For evaluation of visual performance, we recruited 25 volunteers to collect human feedback on the synthetic results as MOS (Mean Opinion Score). Specially, 300 pairs from SHHQ dataset are randomly selected. Each volunteer is asked to select the generated result with the best visual performance from each group of images.

\begin{table}[t]
	\centering
	\scriptsize
	  \caption{Ablation study of SETA. Lower LPIPS indicates better results. Higher MOS indicates that humans prefer.}
 	  \vspace{2mm}
	   	   \setlength{\tabcolsep}{0.3mm}\renewcommand{\arraystretch}{1.5}
	   {
\begin{tabular}{cccccc}
\toprule[1pt]
\multirow{2}{*}{\footnotesize Methods}                & \multicolumn{2}{c}{\footnotesize NTED} &  & \multicolumn{2}{c}{\footnotesize CASD} \\ \cline{2-3} \cline{5-6} 
                                        &\footnotesize LPIPS       & \footnotesize MOS       &  & \footnotesize LPIPS            & \footnotesize  MOS          \\ \hline
 \footnotesize Pre-trained Model                    & \footnotesize 0.211       & \footnotesize 3.34\%      &  & \footnotesize 0.198           & \footnotesize 4.41\%          \\ \hline
  \footnotesize w/ Appearance Adapter            & \footnotesize 0.084       & \footnotesize 20.53\%      &  & \footnotesize 0.079           & \footnotesize 23.43\%          \\ \hline
 \footnotesize w/ Appearance Adapter + Skeleton Adapter  & \footnotesize 0.049       & \footnotesize 76.13\%      &  & \footnotesize 0.042          & \footnotesize 72.17\%          \\ 
\bottomrule[1pt]
\end{tabular}}
\label{tab:ablation_study_detail}
 \vspace{-7mm}
\end{table}

\begin{figure*}[t]
\centering
\includegraphics[width=1\linewidth,height=0.30\linewidth]{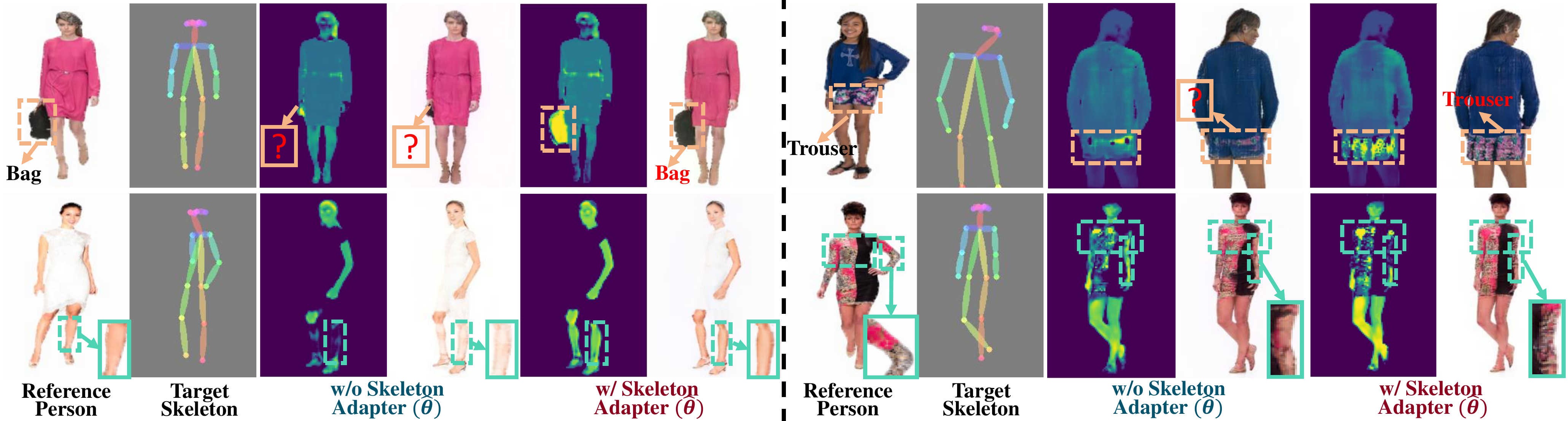}
\vspace{-7mm}
\caption{Qualitative effects of skeleton adapter during pose transformation. Given a reference person image and a target skeleton, we show the feature representations and visualization results of the skeleton adapter. $\hat{\theta}$ denotes the parameters of NTED~\cite{Pose_one_4} has been updated by OOD appearance signals.}
\label{fig:comp_appe_skl}
\vspace{-7mm}
\end{figure*}
\vspace{-4mm}
\paragraph{Effectiveness of Appearance Adapters.} 
The proposed appearance adapter is mainly used to provide a stable appearance transition. (1) Table~\ref{tab:ablation_study_detail} shows quantitative evaluations on SHHQ~\cite{SHHQ} dataset. Appearance adapter could achieve 0.127 LPIPS gains and 20.53\% MOS for NTED~\cite{Pose_one_4}. Similarly, it achieves 0.119 LPIPS gains and 23.43\% MOS for CASD~\cite{Pose_one_3}. The quantitative results shows that using this module is helpful to obtains better realistic results. (2) As illustrated in the second and fifth columns of Fig.~\ref{fig:comp_appe_skl}, the results with appearance adapter perform reasonably well in maintaining the global appearance texture of various OOD data, which is help to train skeleton adapter stably.

\vspace{-3mm}
\paragraph{Effectiveness of Skeleton Adapter.} 
To verify the skeleton adapter comprehensively, we perform it to learn implicit posture representation and make the analyses from \emph{quantitation, human feedback, and visual quality}: (1) As depicted in Table~\ref{tab:ablation_study_detail}, joint training with skeleton adapter can further improve LPIPS performance by 41.66\% for NTED and 46.84\% for CASD. We also collect human feed back for skeleton adapters. The synthesis results achieves nearly 76.13\% MOS for NTED and 72.17\% for CASD. It suggests that skeleton adapter can keep the most useful features and maintain a stable feature-warping process in pose transformation stage. (2) As shown in the first row of Fig.~\ref{fig:comp_appe_skl}, `w/o skeleton adapter ($\hat{\theta}$)' mistakenly ignores some noticeable features, leading to attribute missing and distorted embroideries. In the second row, images generated by `w/o skeleton adapter ($\hat{\theta}$)' exist the clear artifacts, which are caused by erroneous feature warping in pose transition. Generally, only fetched appearance signals from augmented derivatives, the pre-trained model cannot handle highly non-rigid deformation when encounter various OOD skeleton. Therefore, we combine the pre-trained model with skeleton adapter to keep motion consistency. As shown in Fig.~\ref{fig:comp_appe_skl}, `w/ skeleton adapter ($\hat{\theta}$)' could maintain the human features in various OOD skeleton, which is well impressive as none of pose transfer methods have done it before.

\begin{table}[t]
	\centering
	\scriptsize
	  \caption{Analysis for the number of updating iterations in test-time training.}
 	  \vspace{2mm}
	   
	   \setlength{\tabcolsep}{1.3mm}\renewcommand{\arraystretch}{1.5}
	   {
	   
\begin{tabular}{ccccccc}
\toprule[1pt]
\multirow{2}{*}{\begin{tabular}[c]{@{}c@{}}\small Test-time\\ \small Training\end{tabular}} & \multicolumn{2}{c}{\small NTED} & \multicolumn{2}{c}{\small CASD} & \multicolumn{2}{c}{\small ADGAN} \\ \cline{2-7} 
                                                                           & \small LPIPS        & \small Time     & \small LPIPS        & \small Time     & \small LPIPS        &\small Time      \\ \hline
\small w/o SETA                                                                  & \small 0.211      & \small 0.68$s$        & \small 0.198      & \small 1.12$s$       & \small 0.425      & \small 1.36$s$         \\ \hline
\small 5 iterations                                                                 & \small 0.105      & \small 10.87$s$       &  \small 0.096          &  \small 24.8$s$           & \small 0.163      & \small 26.5$s$         \\ \hline
\small 10 iterations                                                                 & \small 0.077      & \small 14.21$s$       & \small 0.073     & \small 30.95$s$       & \small 0.115      & \small 33.08$s$        \\ \hline
\small 30 iterations                                                                 & \small 0.049      & \small 28.48$s$       & \small 0.042           & \small  60.46$s$        & \small 0.079      & \small 61.87$s$      \\ 
\bottomrule[1pt]
\end{tabular}}
\label{tab:time_ablation}
\vspace{-7mm}
\end{table}

\vspace{-4mm}
\paragraph{Discussion of Domain Succession.}
Since we first use two domain information in TTA, we study the sequential relationship between two domains. (1) \emph{Could we fetch both OOD appearance and skeleton signals simultaneously to update the pre-trained model?} Since the pair-wise label of OOD appearance and skeleton are absent, we could not obtain the disentangled signals synchronously. (2) \emph{Could we fetch OOD skeleton signals at first?} Because the pre-trained model could not handle OOD appearance beyond the DeepFashion dataset, it is hard to directly realize the motion consistency between the input image and the generated images. Thus, we recommend to use skeleton adapter to learn implicit posture representation from augmented derivatives at first step, which is beneficial to provide a stable pose transition for learning skeleton signals. (3) \emph{Could we fetch OOD appearance signals at first time?} Inspired by the typical pipelines pose transfer, which first extract the appearance texture and then distribute them to the target skeletons. We first learn OOD appearance signals from OOD data, which exhibits a strong adaptability for the appearance domain. Then, we generate the pose transfer image with OOD appearance, and fetch OOD skeleton signals by using the consistency over motion.  By this means, we not only make the connection between the two domains naturally, but also avoid catastrophic forgetting caused by unstable adaption.

\vspace{-3mm}
\paragraph{Analysis of Test-time Updating Iterations.}

To strike a trade-off between performance and update speed, we conduct various experiments on SHHQ for update iterations in the appearance adaption stage for SETA. The quantitative results of the update iterations are presented in Table~\ref{tab:time_ablation}. It can be observed that larger update iterations yield better generated results, but also bring more time consumption on adaption stage.

\vspace{-2mm}

\section{Conclusion and Limitation}

In this paper, we first extend the pose transfer task to the open-world environment. Specifically, we propose SEquential Test-time Adaption (SETA) to learn non-trivial signals in open-world condition. Extensive evaluations clearly verify the effectiveness of SETA over the state-of-the-art methods with more similar identity, less twisty pattern and greater generalization ability.

Though SETA has learned the non-trivial signals in open-world condition, some detail textures (\emph{e.g.} cloth, makeup) still contain artifacts. In following work, we will apply generative model with ultra high-resolution (UHR) images in our algorithm for a further experiment.


{\small
\normalem
\bibliographystyle{ieee_fullname}
\bibliography{egbib}

\begin{thebibliography}{10}\itemsep=-1pt

\bibitem{Lilun2}
Peshal Agarwal, Danda~Pani Paudel, Jan-Nico Zaech, and Luc Van~Gool.
\newblock Unsupervised robust domain adaptation without source data.
\newblock {\em arXiv preprint arXiv:2103.14577}, 2021.

\bibitem{Pose_multi_1}
Guha Balakrishnan, Amy Zhao, Adrian~V Dalca, Fredo Durand, and John Guttag.
\newblock Synthesizing images of humans in unseen poses.
\newblock In {\em Proceedings of the IEEE conference on computer vision and
  pattern recognition}, pages 8340--8348, 2018.

\bibitem{DG1}
Gilles Blanchard, Gyemin Lee, and Clayton Scott.
\newblock Generalizing from several related classification tasks to a new
  unlabeled sample.
\newblock In {\em NIPS}, 2011.

\bibitem{DG2}
Gilles Blanchard, Gyemin Lee, and Clayton Scott.
\newblock Generalizing from several related classification tasks to a new
  unlabeled sample.
\newblock In {\em NIPS}, 2011.

\bibitem{DG6}
Junbum Cha, Sanghyuk Chun, Kyungjae Lee, Han-Cheol Cho, Seunghyun Park, Yunsung
  Lee, and Sungrae Park.
\newblock Swad: Domain generalization by seeking flat minima.
\newblock In {\em NIPS}, 2021.

\bibitem{Deblurring}
Zhixiang Chi, Yang Wang, Yuanhao Yu, and Jin Tang.
\newblock Test-time fast adaptation for dynamic scene deblurring via
  meta-auxiliary learning.
\newblock In {\em CVPR}, 2021.

\bibitem{DG3}
Xinjie Fan, Qifei Wang, Junjie Ke, Feng Yang, Boqing Gong, and Mingyuan Zhou.
\newblock Adversarially adaptive normalization for single domain
  generalization.
\newblock In {\em CVPR}, 2021.

\bibitem{SHHQ}
Jianglin Fu, Shikai Li, Yuming Jiang, Kwan-Yee Lin, Chen Qian, Chen~Change Loy,
  Wayne Wu, and Ziwei Liu.
\newblock Stylegan-human: A data-centric odyssey of human generation.
\newblock {\em arXiv preprint arXiv:2204.11823}, 2022.

\bibitem{gandelsman2022test}
Yossi Gandelsman, Yu Sun, Xinlei Chen, and Alexei~A Efros.
\newblock Test-time training with masked autoencoders.
\newblock {\em arXiv preprint arXiv:2209.07522}, 2022.

\bibitem{DA3}
Yaroslav Ganin and Victor Lempitsky.
\newblock Unsupervised domain adaptation by backpropagation.
\newblock In {\em ICML}, 2015.

\bibitem{Gram}
Leon~A Gatys, Alexander~S Ecker, and Matthias Bethge.
\newblock Image style transfer using convolutional neural networks.
\newblock In {\em CVPR}, 2016.

\bibitem{DeepFashion2}
Yuying Ge, Ruimao Zhang, Xiaogang Wang, Xiaoou Tang, and Ping Luo.
\newblock Deepfashion2: A versatile benchmark for detection, pose estimation,
  segmentation and re-identification of clothing images.
\newblock In {\em CVPR}, 2019.

\bibitem{RL}
Nicklas Hansen, Rishabh Jangir, Yu Sun, Guillem Aleny{\`a}, Pieter Abbeel,
  Alexei~A Efros, Lerrel Pinto, and Xiaolong Wang.
\newblock Self-supervised policy adaptation during deployment.
\newblock {\em arXiv preprint arXiv:2007.04309}, 2020.

\bibitem{FID}
Martin Heusel, Hubert Ramsauer, Thomas Unterthiner, Bernhard Nessler, and Sepp
  Hochreiter.
\newblock Gans trained by a two time-scale update rule converge to a local nash
  equilibrium.
\newblock {\em NIPS}, 2017.

\bibitem{Tiktok}
Yasamin Jafarian and Hyun~Soo Park.
\newblock Learning high fidelity depths of dressed humans by watching social
  media dance videos.
\newblock In {\em CVPR}, 2021.

\bibitem{VGGloss}
Justin Johnson, Alexandre Alahi, and Li Fei-Fei.
\newblock Perceptual losses for real-time style transfer and super-resolution.
\newblock In {\em ECCV}, 2016.

\bibitem{Lilun4}
Jogendra~Nath Kundu, Naveen Venkat, R~Venkatesh Babu, et~al.
\newblock Universal source-free domain adaptation.
\newblock In {\em CVPR}, 2020.

\bibitem{Parsing}
Peike Li, Yunqiu Xu, Yunchao Wei, and Yi Yang.
\newblock Self-correction for human parsing.
\newblock {\em IEEE Transactions on Pattern Analysis and Machine Intelligence},
  2020.

\bibitem{Lilun3}
Rui Li, Qianfen Jiao, Wenming Cao, Hau-San Wong, and Si Wu.
\newblock Model adaptation: Unsupervised domain adaptation without source data.
\newblock In {\em CVPR}, 2020.

\bibitem{Pose_multi_4}
Yining Li, Chen Huang, and Chen~Change Loy.
\newblock Dense intrinsic appearance flow for human pose transfer.
\newblock In {\em CVPR}, 2019.

\bibitem{Lilun1}
Jian Liang, Dapeng Hu, and Jiashi Feng.
\newblock Do we really need to access the source data? source hypothesis
  transfer for unsupervised domain adaptation.
\newblock In {\em ICML}, 2020.

\bibitem{Dehazing}
Huan Liu, Zijun Wu, Liangyan Li, Sadaf Salehkalaibar, Jun Chen, and Keyan Wang.
\newblock Towards multi-domain single image dehazing via test-time training.
\newblock In {\em CVPR}, 2022.

\bibitem{Pose_one_6}
Songhua Liu, Jingwen Ye, Sucheng Ren, and Xinchao Wang.
\newblock Dynast: Dynamic sparse transformer for exemplar-guided image
  generation.
\newblock {\em arXiv preprint arXiv:2207.06124}, 2022.

\bibitem{liu2016deepfashion}
Ziwei Liu, Ping Luo, Shi Qiu, Xiaogang Wang, and Xiaoou Tang.
\newblock Deepfashion: Powering robust clothes recognition and retrieval with
  rich annotations.
\newblock In {\em CVPR}, 2016.

\bibitem{DA5}
Ziwei Liu, Zhongqi Miao, Xingang Pan, Xiaohang Zhan, Dahua Lin, Stella~X Yu,
  and Boqing Gong.
\newblock Open compound domain adaptation.
\newblock In {\em CVPR}, 2020.

\bibitem{DA4}
Mingsheng Long, Yue Cao, Jianmin Wang, and Michael Jordan.
\newblock Learning transferable features with deep adaptation networks.
\newblock In {\em ICML}, 2015.

\bibitem{Pose_multi_3}
Zhengyao Lv, Xiaoming Li, Xin Li, Fu Li, Tianwei Lin, Dongliang He, and
  Wangmeng Zuo.
\newblock Learning semantic person image generation by region-adaptive
  normalization.
\newblock In {\em CVPR}, 2021.

\bibitem{Pose_1}
Liqian Ma, Xu Jia, Qianru Sun, Bernt Schiele, Tinne Tuytelaars, and Luc
  Van~Gool.
\newblock Pose guided person image generation.
\newblock {\em NIPS}, 2017.

\bibitem{Pose_one_2}
Yifang Men, Yiming Mao, Yuning Jiang, Wei-Ying Ma, and Zhouhui Lian.
\newblock Controllable person image synthesis with attribute-decomposed gan.
\newblock In {\em CVPR}, 2020.

\bibitem{DA10}
Krikamol Muandet, David Balduzzi, and Bernhard Sch{\"o}lkopf.
\newblock Domain generalization via invariant feature representation.
\newblock In {\em ICML}, 2013.

\bibitem{DG4}
Prashant Pandey, Mrigank Raman, Sumanth Varambally, and Prathosh AP.
\newblock Domain generalization via inference-time label-preserving target
  projections.
\newblock In {\em CVPR}, 2021.

\bibitem{Pose_one_4}
Yurui Ren, Xiaoqing Fan, Ge Li, Shan Liu, and Thomas~H Li.
\newblock Neural texture extraction and distribution for controllable person
  image synthesis.
\newblock In {\em CVPR}, 2022.

\bibitem{Pose_multi_2}
Yurui Ren, Xiaoming Yu, Junming Chen, Thomas~H Li, and Ge Li.
\newblock Deep image spatial transformation for person image generation.
\newblock In {\em CVPR}, 2020.

\bibitem{DA1}
Kate Saenko, Brian Kulis, Mario Fritz, and Trevor Darrell.
\newblock Adapting visual category models to new domains.
\newblock In {\em ECCV}, 2010.

\bibitem{DA2}
Kuniaki Saito, Kohei Watanabe, Yoshitaka Ushiku, and Tatsuya Harada.
\newblock Maximum classifier discrepancy for unsupervised domain adaptation.
\newblock In {\em CVPR}, 2018.

\bibitem{MM_TTA}
Inkyu Shin, Yi-Hsuan Tsai, Bingbing Zhuang, Samuel Schulter, Buyu Liu, Sparsh
  Garg, In~So Kweon, and Kuk-Jin Yoon.
\newblock Mm-tta: Multi-modal test-time adaptation for 3d semantic
  segmentation.
\newblock In {\em CVPR}, 2022.

\bibitem{Super_resolution}
Assaf Shocher, Nadav Cohen, and Michal Irani.
\newblock “zero-shot” super-resolution using deep internal learning.
\newblock In {\em CVPR}, 2018.

\bibitem{DG5}
Yang Shu, Zhangjie Cao, Chenyu Wang, Jianmin Wang, and Mingsheng Long.
\newblock Open domain generalization with domain-augmented meta-learning.
\newblock In {\em CVPR}, 2021.

\bibitem{TTT}
Yu Sun, Xiaolong Wang, Zhuang Liu, John Miller, Alexei Efros, and Moritz Hardt.
\newblock Test-time training with self-supervision for generalization under
  distribution shifts.
\newblock In {\em ICML}, 2020.

\bibitem{ReID_model}
Yifan Sun, Liang Zheng, Yi Yang, Qi Tian, and Shengjin Wang.
\newblock Beyond part models: Person retrieval with refined part pooling (and a
  strong convolutional baseline).
\newblock In {\em ECCV}, 2018.

\bibitem{DA11}
Riccardo Volpi, Diane Larlus, and Gr{\'e}gory Rogez.
\newblock Continual adaptation of visual representations via domain
  randomization and meta-learning.
\newblock In {\em CVPR}, 2021.

\bibitem{Tent}
Dequan Wang, Evan Shelhamer, Shaoteng Liu, Bruno Olshausen, and Trevor Darrell.
\newblock Tent: Fully test-time adaptation by entropy minimization.
\newblock In {\em ICLR}, 2021.

\bibitem{ReID_loss}
Jian Wang, Yunshan Zhong, Yachun Li, Chi Zhang, and Yichen Wei.
\newblock Re-identification supervised texture generation.
\newblock In {\em CVPR}, 2019.

\bibitem{SSIM}
Zhou Wang, Alan~C Bovik, Hamid~R Sheikh, and Eero~P Simoncelli.
\newblock Image quality assessment: from error visibility to structural
  similarity.
\newblock {\em IEEE transactions on image processing}, 13(4):600--612, 2004.

\bibitem{DA12}
Xiangyu Yue, Yang Zhang, Sicheng Zhao, Alberto Sangiovanni-Vincentelli, Kurt
  Keutzer, and Boqing Gong.
\newblock Domain randomization and pyramid consistency: Simulation-to-real
  generalization without accessing target domain data.
\newblock In {\em ICCV}, 2019.

\bibitem{Pose_one_5}
Pengze Zhang, Lingxiao Yang, Jian-Huang Lai, and Xiaohua Xie.
\newblock Exploring dual-task correlation for pose guided person image
  generation.
\newblock In {\em CVPR}, 2022.

\bibitem{Lpips}
Richard Zhang, Phillip Isola, Alexei~A Efros, Eli Shechtman, and Oliver Wang.
\newblock The unreasonable effectiveness of deep features as a perceptual
  metric.
\newblock In {\em CVPR}, 2018.

\bibitem{market_1501}
Liang Zheng, Liyue Shen, Lu Tian, Shengjin Wang, Jingdong Wang, and Qi Tian.
\newblock Scalable person re-identification: A benchmark.
\newblock In {\em ICCV}, 2015.

\bibitem{Pose_one_3}
Xinyue Zhou, Mingyu Yin, Xinyuan Chen, Li Sun, Changxin Gao, and Qingli Li.
\newblock Cross attention based style distribution for controllable person
  image synthesis.
\newblock {\em arXiv preprint arXiv:2208.00712}, 2022.

\bibitem{Pose_one_1}
Zhen Zhu, Tengteng Huang, Baoguang Shi, Miao Yu, Bofei Wang, and Xiang Bai.
\newblock Progressive pose attention transfer for person image generation.
\newblock In {\em CVPR}, 2019.

\end{thebibliography}
}

\appendix

\newpage




\begin{figure*}[h]
	\begin{center}
		\includegraphics[width=1\linewidth]{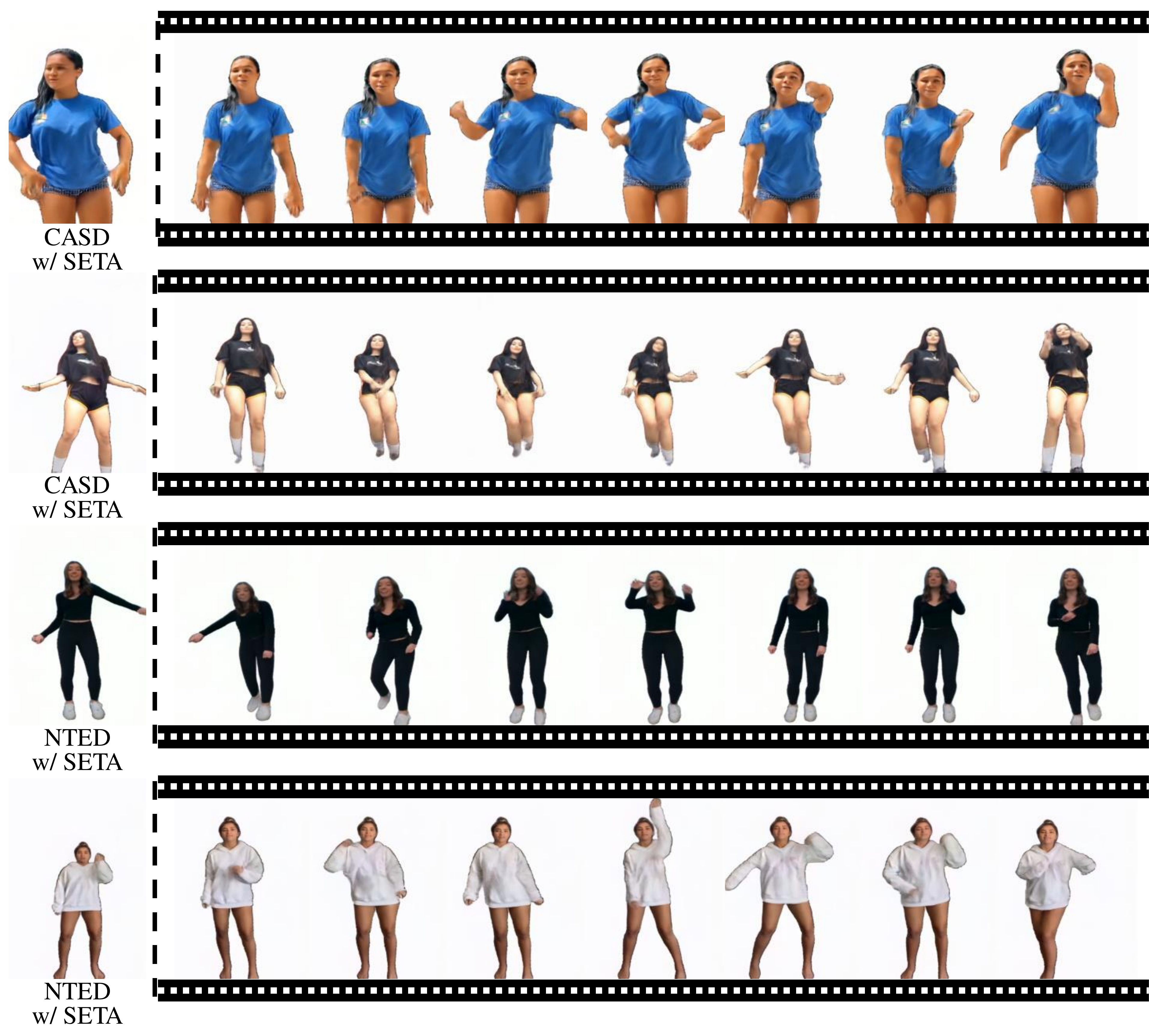}
	\end{center}
	\caption{Skeleton-driven results on the Tiktok dataset~\cite{Tiktok}. Given target skeletons, our method can generate realistic dance sequences.}
	\label{fig:video_comp}
\end{figure*}


\begin{figure*}[t]
	\begin{center}
		\includegraphics[width=1\linewidth]{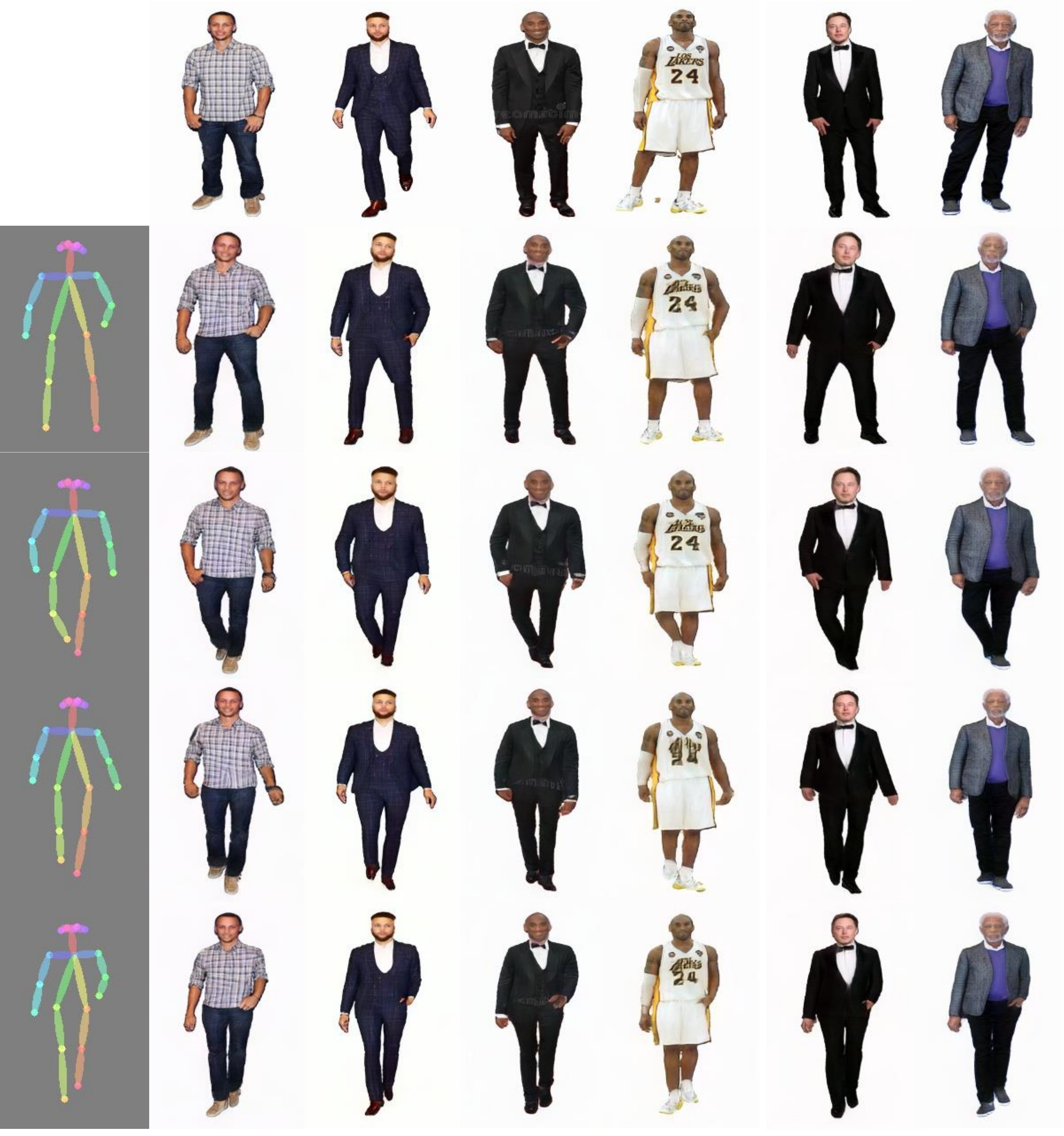}
	\end{center}
    \caption{Examples of celebrity motion synthesis. Our algorithm transfer Stephen Curry,  Kobe Bean Bryant, Elon Reeve Musk and Morgan Freeman into desired postures.}
	\label{fig:celebrity_1}
\end{figure*}
\newpage
\begin{figure*}[t]
	\begin{center}
		\includegraphics[width=1\linewidth]{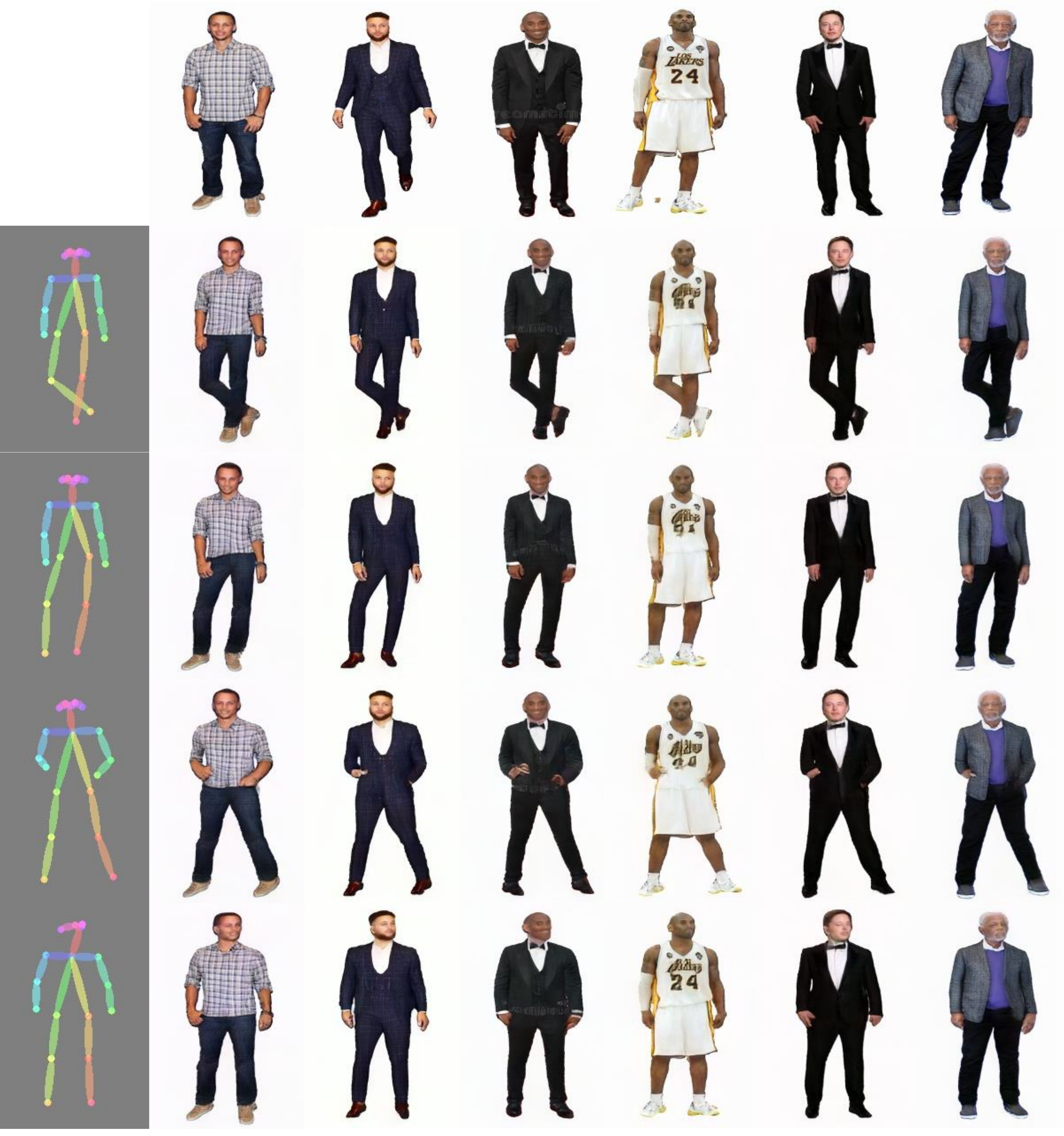}
	\end{center}
    \caption{Examples of celebrity motion synthesis. Our algorithm transfer Stephen Curry,  Kobe Bean Bryant, Elon Reeve Musk and Morgan Freeman into desired postures.}
	\label{fig:celebrity_2}
\end{figure*}

\end{document}